\documentclass[10pt,twocolumn,letterpaper]{article}

\usepackage{cvpr}
\usepackage{times}
\usepackage{epsfig}
\usepackage{graphicx}
\usepackage{amsmath}
\usepackage{amssymb}

\usepackage[breaklinks=true,bookmarks=false]{hyperref}

\usepackage{framed,multirow}
\usepackage{hyperref}

\cvprfinalcopy 

\def\cvprPaperID{****} 

\begin{document}

\title{ASSD: Attentive Single Shot Multibox Detector}

\author{Jingru Yi,\quad Pengxiang Wu,\quad Dimitris N. Metaxas\\
Department of Computer Science, Rutgers University,
Piscataway, NJ 08854, USA\\
{\tt\small jy486,pw241,dnm@cs.rutgers.edu}
}

\maketitle

\begin{abstract}
This paper proposes a new deep neural network for object detection. The proposed network, termed ASSD, builds feature relations in the spatial space of the feature map. With the global relation information, ASSD learns to highlight useful regions on the feature maps while suppressing the irrelevant information, thereby providing reliable guidance for object detection. Compared to methods that rely on complicated CNN layers to refine the feature maps, ASSD is simple in design and is computationally efficient. Experimental results show that ASSD competes favorably with the state-of-the-arts, including SSD, DSSD, FSSD and RetinaNet. Code is available at: \url{https://github.com/yijingru/ASSD-Pytorch}.
\end{abstract}

\section{Introduction}
\label{sec:intro}
In recent years, object detection has experienced a rapid development with the aid of convolutional neural networks (CNN). Generally, the CNN-based object detectors can be divided into two types: one-stage object detector and two-stage object detector. The two-stage object detectors, such as R-CNN \cite{girshick2014rich}, Fast and Faster R-CNN \cite{Girshick_2015_ICCV,ren2015faster} and SPPnet \cite{he2014spatial}, are proposal driven, with a second stage for refining the detection. However, these two-stage object detectors are inefficient for real-time applications due to the decoupled multi-stage processing. In contrast, the one-stage object detectors, including YOLO \cite{redmon2016you}, YOLO-v2 \cite{redmon2017yolo9000} and SSD \cite{liu2016ssd}, propose to model the object detection as a simple regression problem and encapsulate all the computation in a single feed-forward CNN, thereby speeding up the detection to a large extent. However, the one-stage detectors are generally less accurate than the two-stage ones. The main reason would be the extreme foreground-background class imbalance of the dense anchor boxes \cite{lin2018focal}. To solve this issue, RetinaNet \cite{lin2018focal} proposes a focal loss to train its FPN-based \cite{lin2017feature} one-stage detector. However, the focal loss is parameter sensitive, and it would require exhaustive experiments to obtain the optimal parameters.

\begin{figure}[!ht]
	\centering
	\includegraphics[width=0.43\textwidth]{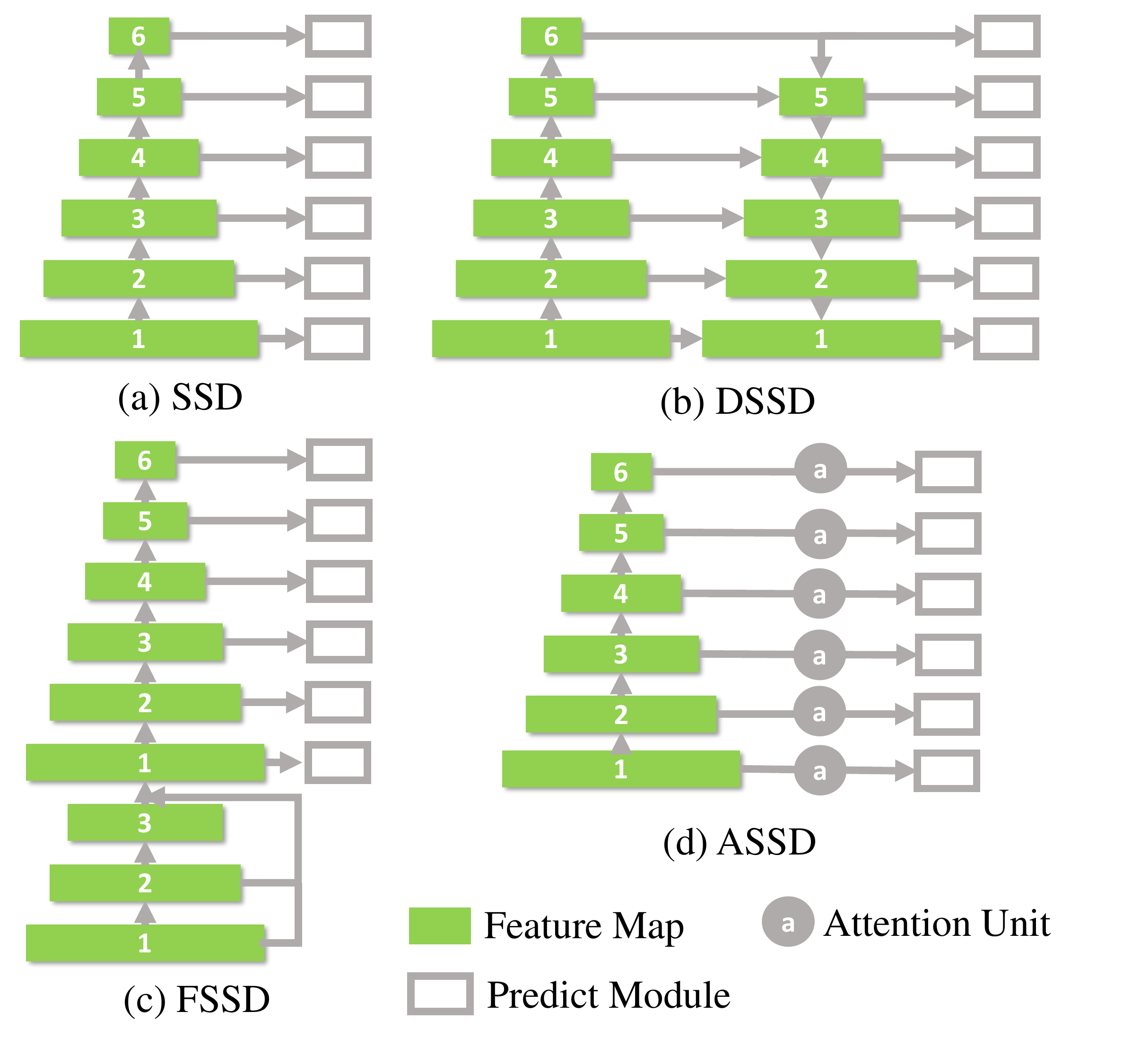}
	\caption{The structures of different SSD-based detectors. (a) SSD \cite{liu2016ssd}, (b) DSSD \cite{fu2017dssd}, (c) FSSD \cite{li2017fssd}, (d) ASSD (Ours).}
	\label{fig:fig1}
\end{figure}

In this paper, we aim to improve the one-stage detectors from a different perspective. We propose to discover the intrinsic feature relations on the feature map to focus the detector on regions that are critical to the detection task. Our key motivation comes from the human vision system. When perceiving a scene, humans first glance at the scene and then instantly figure out the contents through global dependency analysis. Besides, when the eyeballs focus on a fixation point, the resolution of the neighboring regions decreases. To simulate such human vision mechanism, we design an \textit{attention unit} that is capable of analyzing the importance of features at different positions, based on the global feature relations. The attention unit is fully differentiable and in-place. This design generates the \textit{attention maps} which highlight the useful regions and suppress the irrelevant information. Compared to methods that only build relations among proposals \cite{Hu_2018_CVPR,zeng2018crafting}, our method considers the global feature correlations at pixel level and conforms to the visual mechanism of humans.

We choose the SSD as our base one-stage detector, which provides the optimal trade-off among simplicity, speed and accuracy. Combined with the attention unit, we term the resulting object detector as Attentive SSD (ASSD). ASSD is simpler in design and more effective at refining the contextual semantics compared to the existing SSD-based detectors (see Fig.~\ref{fig:fig1}). In particular, DSSD \cite{fu2017dssd} relies on a complex feature pyramid to encourage the information flow among different layers. While achieving better accuracies than the original SSD, it is relatively more complex and thus computationally inefficient. Another recent approach, FSSD \cite{li2017fssd}, builds additional fusion modules for multi-scale feature aggregation, but only achieves marginal improvements upon SSD. In contrast to these works, our ASSD retains the original structure of SSD and employs a single efficient attention unit to refine the object information from each layer (see Fig.~\ref{fig:fig1}d). This design preserves the advantages of the original SSD while being more effective at learning object features.
We demonstrate the advantages of ASSD on a number of representative benchmark datasets, including PASCAL VOC \cite{Everingham15} and COCO \cite{lin2014microsoft}. Experimental results validate the superiority of ASSD compared to the state-of-the-arts in terms accuracy and efficiency.
Our main contributions can be summarized as follows:
\begin{enumerate}
\item We propose to incorporate pixel-wise feature relations into the one-stage detector. Our design follows the human vision mechanism and facilitates the object feature learning. 
\item The proposed network preserves the simplicity and efficiency of SSD while being more accurate.
\item We perform a series of experiments to validate the advantages of ASSD. The experimental results show that ASSD competes favorably with the state-of-the-arts in terms of accuracy and efficiency.
\end{enumerate}

\section{Related Works}
\label{sec:related_works}

\begin{figure*}[!tb]
	\centering
	\includegraphics[width=0.8\textwidth]{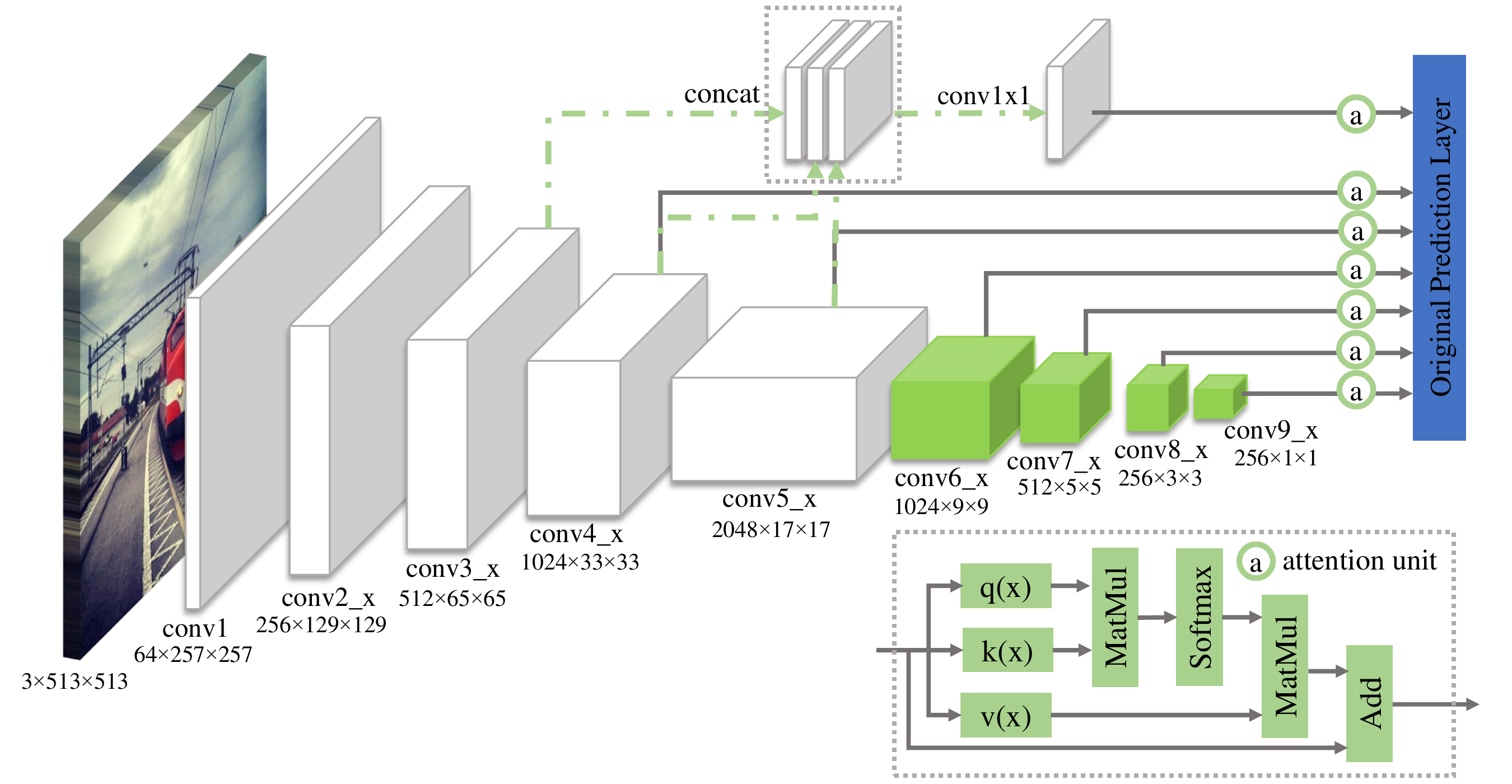}
	\caption{Overview of the ASSD architecture. The backbone of ASSD (conv1-5) is ResNet101 \cite{he2016deep}. The extra convolutional blocks follow the same settings as the original SSD \cite{liu2016ssd}. Batch normalization and ReLU are used in all layers. The feature maps are displayed as ``number of channels $\times$ height $\times$ width". Feature map from conv3 is enhanced by fusion of feature maps from conv3-5.}
	\label{fig:fig2}
\end{figure*}

\subsection{Object Detection}
Object detection involves localization and classification. From traditional hand-crafted feature-based methods (e.g., SIFT \cite{sermanet2013pedestrian} and HOG \cite{dalal2005histograms}) to recent CNN-based models, last decades have witnessed a significant development of object detection techniques. In recent years, CNN-based object detectors have gained remarkable success and generally can be divided into two categories: the proposal-driven two-stage detectors, and the regression-oriented one-stage detectors. 

The two-stage object detectors are composed of two decoupled operations: proposal generation and box refinement. The pioneering work, R-CNN \cite{girshick2014rich}, utilizes selective search to generate region proposals and classifies them with class-specific linear SVM using the learned CNN features. The major weakness of R-CNN is that it needs to perform the forward pass for each proposal, leading to an extremely inefficient model. To solve this issue, SPPnet \cite{he2014spatial} suggests sharing the CNN computation for all proposals, whereas Fast R-CNN \cite{Girshick_2015_ICCV} replaces the SVM with fully-connected layers (FCs) to enable single-stage training without additional feature caching. Faster R-CNN \cite{ren2015faster} goes a step further and introduces a region proposal network (RPN) where the proposal computation is performed through shared CNN features, thereby largely speeding up the detection process. In a more aggressive manner, R-FCN \cite{dai2016r} replaces the FCs with position-sensitive score maps and encodes translation variance information into these maps, leading to a variance insensitive fully convolutional network (FCN) for accurate object detection. Another recent work, FPN \cite{lin2017feature}, employs a top-down pyramid structure to reuse the higher-resolution features maps from the feature hierarchy and has achieved the state-of-the-art results. Two-stage object detectors are quite effective at object feature learning. However, they are generally inefficient in computation.

Different from two-stage detectors, one-stage object detectors discard the region proposal stage, thereby making the detection more efficient. YOLO \cite{redmon2016you} proposes to use a single CNN to simultaneously predict multiple bounding boxes as well as their class probabilities. While being extremely fast, YOLO is far less accurate than the two-stage models. Instead of directly predicting the coordinates of bounding boxes, YOLOv2 \cite{redmon2017yolo9000} employs the anchor boxes to facilitate the detection and improves the accuracy a lot. From a different perspective, SSD \cite{liu2016ssd} builds a pyramid CNN network on top of the backbone, and detects objects of different scales from the multi-scale feature maps in a single forward pass. SSD has achieved better performance than YOLOv2. Based on SSD and similar to FPN, DSSD \cite{fu2017dssd} employs top-down pyramid CNN layers to improve the accuracy but at the cost of computational efficiency. FSSD \cite{li2017fssd} inserts a fusion module at the bottom of the feature pyramid to enhance the accuracy of SSD. While still being fast, FSSD only achieves marginal improvements upon SSD in accuracy. Other works, such as RefineDet \cite{zhang2018single}, DSOD \cite{shen2017dsod} and STOD \cite{zhou2018scale}, also improve the detection accuracy of SSD either through refinining the anchors or by aggregating the feature maps at different scales. CornerNet \cite{law2018cornernet} follows a different strategy and improves the detection accuracy with keypoint-based object detectors. The recent work, RetinaNet \cite{lin2018focal}, builds the one-stage detector based on FPN and proposes a focal loss for better training. RetinaNet is efficient in inference; however, it requires a large effort for loss function parameter tuning. In this work, we show that by explicitly modeling the feature relations, our ASSD model competes favorably with RetinaNet without heavy tuning of parameters. 

\subsection{Visual Attention}
Visual attention mechanism is generally used to exploit the salient visual information and facilitate visual tasks such as object recognition. There are many visual attention methods in the literature. For example, the saliency-based visual attention model \cite{itti1998model} selects attended locations from saliency maps. In contrast, RAM \cite{mnih2014recurrent}, AttentionNet \cite{yoo2015attentionnet}  and RA-CNN \cite{fu2017look}  search and crop the useful regions recurrently. In particular, RAM employs Recurrent Neural Network (RNN) and reinforcement learning to discover the target. AttentionNet explores the direction that leads to the real object through CNN classification. RA-CNN also uses reinforcement learning to learn the discriminative region attention and region-based feature representation. The common characteristic of these methods is that they only focus on single instance problems. For multi-object recognition, AC-CNN \cite{li2017attentive}, LPA \cite{jetley2018learn} and RelationNet \cite{Hu_2018_CVPR} have been proposed to discover a global contextual guidance. AC-CNN examines the global context through the stacked Long Short-Term Memory (LSTM) units. LPA learns the attention maps from the compatibility scores between the shallow and deep layers. RelationNet correlates the geometry features and appearance information between proposals to generate and forward the attentive features, and it is designed specifically for the two-stage object detectors. In practice, RelationNet only achieves a slight improvement.

\subsection{Self-Attention}
The self-attention mechanism has been widely used in natural language processing (NLP) field to model long-range dependencies of a sentence. LSTMN \cite{cheng2016long}  develops an attention memory network that discovers the relations between tokens to enhance the memorization capability of LSTM. Structured self-attentive sentence embedding \cite{lin2017structured}  introduces self-attention in the bidirectional LSTM to generate a 2-D matrix representation of the embeddings, where each row attends to a different part of the sentence. Transformer \cite{vaswani2017attention} draws global dependencies between input and output based solely on attention mechanisms. Inspired by Transformer, in this work we build the long-range dependencies among all feature pixels within the feature map itself. In a similar spirit to Transformer, our ASSD is capable of attending to different regions for more effective object detection.

\section{Attentive SSD}
\label{sec:method}
SSD \cite{liu2016ssd} performs the detection on multi-scale feature maps to handle various object sizes effectively. However, the shallow layer lacks semantic information and is therefore insufficient for detecting small objects. One way to solving this problem is to build more CNN layers to make further refinements of the feature maps or inject semantics from deep layers to the shallow ones exhaustively. Considering that speed is the key advantage of one-stage object detectors, we aim to improve the SSD accuracy with small extra computational cost. To this end, we construct a small network, namely attention unit, and embed it into SSD to improve the detection accuracy. Our ASSD network architecture is illustrated in Fig.~\ref{fig:fig2}. Specifically, we use ResNet101 (conv1-5) \cite{he2016deep} as the backbone. The pyramid convolutional blocks (conv6-9) follow the same design as the original SSD \cite{liu2016ssd}.  The feature maps from conv3-9 are used to detect objects with different scales.  ASSD places the attention unit between the feature map and the prediction module, where the box regression and object classification are performed. 

\begin{table}[!t]
  \renewcommand{\arraystretch}{1.25}
  \footnotesize
  \caption{Architecture of ASSD with ResNet101 backbone. ReLU and Batch Normalization are used in hidden layers. The input size is 513$\times$ 513.
  }
  \label{tab:Parameter}
  \centering
  \begin{tabular}{lcc}
  \hline
    Layer Name& Output Size & Specifications\\
  \hline
    \rule{0pt}{10pt}
	conv1 	
    \rule[-0pt]{0pt}{10pt}
	& 256$\times$256 & 7$\times$7, 64, stride 2 \\
  \hline
    conv2\_x
    & 128$\times$128 &
    $\begin{bmatrix}
        1\times1, 64 \\
		3\times3, 64 \\
		1\times1, 256
    \end{bmatrix}\times 3$\\
  \hline		
    conv3\_x
	& 64$\times$64 &
    $\begin{bmatrix}
        1\times1, 128 \\
        3\times3, 128 \\
        1\times1, 512
    \end{bmatrix}\times 4$\\
  \hline		
    conv4\_x
    & 32$\times$32 &
    $\begin{bmatrix}
        1\times1, 256 \\
		3\times3, 256 \\
		1\times1, 1024
    \end{bmatrix}\times 23$\\
  \hline
    conv5\_x
    & 8$\times$8 &
    $\begin{bmatrix}
        1\times1, 512 \\
		3\times3, 512 \\
		1\times1, 2048
    \end{bmatrix}\times 3$\\
  \hline
  \end{tabular}
\end{table}

\begin{table*}[!t]
    \caption{Comparison of speed and accuracy on PASCAL VOC2007 test. 07+12: 07 trainval+12 trainval. We compared our ASSD with Faster R-CNN \cite{ren2015faster,he2016deep}, R-FCN \cite{dai2016r},YOLOv2 \cite{redmon2017yolo9000},  SSD300*, SSD512*  \cite{liu2016ssd}, SSD321, SSD513, DSSD321, DSSD513 \cite{fu2017dssd}, FSSD300, FSSD513  \cite{li2017fssd}, RefineDet \cite{zhang2018single}. Att is the abbreviation for attention module.}
    \centering
    \begin{tabular}{ | l | c | c | c | c | c | c | c | c| }
      \hline	
      Method & Backbone & Training Data& mAP  &Input Size & FPS & GPU & \#Anchors & \#Parameters\\ \hline
      Faster R-CNN & VGG16& 07+12 &73.2&$\sim$1000$\times$600& 7 &Titan X&6000 & 134.7M\\
      Faster R-CNN & ResNet101 & 07+12 &76.4& $\sim$1000$\times$600&2.4 &K40 &300 & -\\
      R-FCN & ResNet101 & 07+12& 79.5&$\sim$1000$\times$600& 9 &Titan X &300 & 50.9M\\
      YOLOv2  &Darknet19&07+12&78.6&544$\times$544&40&Titan X&- & -\\
      \hline
      RetinaNet300& ResNet101& 07+12& 62.9& 300$\times$300& 11.4&  K80& 15354 &55.7M
      \\
      RetinaNet300+att& ResNet101& 07+12& 64.9& 300$\times$300& 11.1&  K80& 15354 &55.8M
      \\
      SSD300*& VGG16& 07+12&77.5&300$\times$300&46& Titan X&8732 & -\\
      SSD321& ResNet101& 07+12&77.1&321$\times$321& 11.2& Titan X&17080 & 56.8M\\
      FSSD300 & VGG16& 07+12&78.8& 300$\times$300& 65.8& 1080Ti& 8732 & -\\
      DSSD321& ResNet101& 07+12&78.6&321$\times$321&9.5& Titan X &17080 & -  
      \\ 
      ASSD300& VGG16& 07+12& 80.0&300$\times$300& 11.8&K40 &8732   &29.4M
      \\
      ASSD321& ResNet101& 07+12&79.5&321$\times$321&27.5/11.4&Titan X/K40&10325  &66.7M
      \\
      RefineDet320 & VGG16 & 07+12 & 79.5&320$\times$320 &12.9&K80&6375 &32.1M
      \\
      RefineDet320+att & VGG16 & 07+12 & 80.0&320$\times$320 &12.0&K80&6375 &33.9M
      \\
      \hline
      RetinaNet500& ResNet101& 07+12& 72.2& 500$\times$500& 7.1&  K80& 35964&55.7M
      \\
      RetinaNet500+att& ResNet101& 07+12& 73.4& 500$\times$500& 6.7&  K80& 35964 &55.8M
      \\
      SSD512*& VGG16& 07+12&79.5&512$\times$512&19& Titan X&24564 & -\\ 
      SSD513  & ResNet101& 07+12  &80.6& 513$\times$513&6.8& Titan X&43688 & 57.5M \\ 
      FSSD513 & VGG16 & 07+12  &80.9&512$\times$512&35.7& 1080Ti &24564  & - \\ 
      DSSD513 & ResNet101 & 07+12&81.5&513$\times$513 &  5.5& Titan X &43688 & -  \\  
      ASSD512 & VGG16 & 07+12&81.6 & 512$\times$512   & 3.4& K40 &24564&30.2M
      \\ 
      ASSD513 & ResNet101 & 07+12&\textbf{83.0}& 513$\times$513 & 16.0/6.1&Titan X/K40&25844 &67.5M
      \\  
      RefineDet512 & VGG16 & 07+12 & 81.2&512$\times$512 &5.6&K80&16320&32.1M
      \\
      RefineDet512+att & VGG16 & 07+12 & 82.2&512$\times$512 &5.0&K80&16320 &33.9M
      \\
      \hline 
    \end{tabular}
    \label{tab:VOC2007}
\end{table*}

\subsection{Attention Unit}
We adapt the self-attention mechanism from the sequence transduction problem \cite{vaswani2017attention} to our task. In sequence transduction, self-attention mechanism draws global dependencies between the input and output sequences by an attention function, which maps a query and a set of key-value pairs to an output. In self-attention, the attention is motivated by the input features and used for refining these features. Here we repurpose our problem as a similar query problem that estimates the relevant information from the input features in order to build global pixel-level feature correlations.

Suppose $ \mathbf{x^s} \in \mathbb{R}^{C^s\times N^s} $ is the feature map at a given scale $ s\in \{1,\cdots,S\} $, with $ C $ and $ N $ representing the number of channels and total spatial locations in the feature map, respectively. We first linearly transform the feature map $ \mathbf{x^s} $ into three different feature spaces $ \mathbf{q}, \mathbf{k} $ and $ \mathbf{v} $, i.e.,
$ \mathbf{q}(\mathbf{x^s}) = \mathbf{W_q^s}^{\top}\mathbf{x^s} $,
$ \mathbf{k}(\mathbf{x^s}) = \mathbf{W_k^s}^{\top}\mathbf{x^s} $, and
$ \mathbf{v}(\mathbf{x^s}) = \mathbf{W_v^s}^\top\mathbf{x^s} $, where $\mathbf{W_q^s},\mathbf{W_k^s}\in \mathbb{R}^{C^s\times C^\prime} $ and $ \mathbf{W_v^s}\in \mathbb{R}^{C^s\times C^s} $ with $C^\prime = C^s/8$. The attention score matrix $ \mathbf{a^s}\in \mathbb{R}^{N^s\times N^s} $ is then calculated by the matrix multiplication of $ \mathbf{q}(\mathbf{x^s}) $ and $ \mathbf{k}(\mathbf{x^s}) $, as shown in Fig.~\ref{fig:fig2}. Each row of the attention score matrix is normalized by a softmax operation:
\begin{equation} \label{eq1}
\begin{split}
\bar{a}^s_{ij}&= \frac{\exp(a^s_{ij})}{\sum_{j}^{N^s}\exp(a^s_{ij})}, i,j=1,2,\cdots,N^s, \\
\mathbf{a^s} &= \mathbf{q(x^s)}^\top\mathbf{k(x^s)},
\end{split}
\end{equation}
where $ \mathbf{\bar{a}^s_i} $ describes the pixel relations when querying the $ i $-th location of the feature map. We call $ \mathbf{\bar{a}^s_i} $ as an attention map. Note that, the reason we transform the input feature $\mathbf{x^s}$ into $\mathbf{q}$ and $\mathbf{k}$ is to reduce computational cost. The matrix computation of $ \mathbf{q}(\mathbf{x^s}) $ and $ \mathbf{k}(\mathbf{x^s}) $ calculates the feature similarities and creates an $N\times N$ attention map that reveals the feature relations. Note that such pixel-wise relations are learned through the network. 

Next, we apply a matrix multiplication between $ \mathbf{v}(\mathbf{x^s}) $ and the attention maps $ \mathbf{\bar{a}^s} $. In this way we compute an updated feature map as the weighted sums of individual features at each location. Finally, we add the matrix multiplication result back to the input feature map $ \mathbf{x^s} $:
\begin{equation}\label{eq2}
\mathbf{x^{s^\prime}} = \mathbf{x^{s}}+(\mathbf{\bar{a}^s}\mathbf{v}(\mathbf{x^s})^\top)^{\top}.
\end{equation}
Attention map $ \mathbf{\bar{a}^s} $ relates the long-range dependencies of features at all positions and therefore learns global contexts of the feature map. It highlights the relevant parts of the feature map and guides the detection with refined information.

\begin{table}[t!]
    \centering
    \caption{Ablation Study on PASCAL VOC2007 test dataset. Training dataset is 07+12: 07 trainval+12 trainval. Time is evaluated on a single NVIDIA K40 GPU. Note that SSD513+fusion is different from FSSD513 \cite{li2017fssd}.}
    \begin{tabular}{|c|c|c|c|}
      \hline	
      Method & Backbone & Time (s)& mAP \\ \hline
      SSD513 & ResNet101 & 0.1417 & 79.75 \\   
      SSD513+fusion & ResNet101 & 0.1466 & 79.57 \\   
      SSD513+att & ResNet101 & 0.1593 & 82.13 \\    
      SSD513+fusion+att & ResNet101 & 0.1648 & \textbf{82.95} \\  \hline
    \end{tabular}
    \label{tab:Ablation}
\end{table}

\begin{table*}[h!]
	\caption{PASCAL VOC2012 test detection results. Training dataset is 07++12: 07 trainval+07 test+12 trainval. Results are evaluated by online PASCAL VOC evaluation server. We compared the performance of our ASSD321 and ASSD513 with AC-CNN \cite{li2017attentive}, Faster R-CNN \cite{he2016deep}, R-FCN \cite{dai2016r},YOLOv2 \cite{redmon2017yolo9000}, SSD300*, SSD512* \cite{liu2016ssd}, SSD321, SSD513, DSSD321, DSSD 513 \cite{fu2017dssd}, RefineDet \cite{zhang2018single}. }
	\centering
	\resizebox{\textwidth}{!}{%
		\setlength{\tabcolsep}{1.75pt}
		\begin{tabular}{|l|c|c|cccccccccccccccccccc|}
			\hline
			Method & Backbone & mAP & aero & bike & bird & boat & bottle & bus & car & cat & chair & cow & table & dog & horse & mbike & person & plant & sheep & sofa & train & tv \\ \hline
			AC-CNN  & VGG16&70.6&83.2&80.8&70.8&54.9&42.1&79.1&73.4&89.7&47.0&75.9&61.8&87.8&80.9&81.8&74.4&37.8&71.6&67.7&83.1&67.4
			\\ 
			Faster&ResNet101&73.8&86.5&81.6&77.2&58.0&51.0&78.6&76.6&93.2&48.6&80.4&59.0&92.1&85.3&84.8&80.7&48.1&77.3&66.5&84.7&65.6\\ 
			R-FCN&ResNet101&77.6&86.9&83.4&81.5&63.8&\textbf{62.4}&81.6&81.1&93.1&58.0&83.8&60.8&92.7&86.0&84.6&84.4&59.0&80.8&68.6&86.1&72.9\\ 
			YOLOv2&Darknet19&73.4&86.3&82.0&74.8&59.2&51.8&79.8&76.5&90.6&52.1&78.2&58.5&89.3&82.5&83.4&81.3&49.1&77.2&62.4&83.8&68.7\\
			\hline
			Retina300&ResNet101&59.8&73.9&68.2&65.1&43.6&32.3&67.2&58.8&83.0&39.6&58.4&4.6&80.9&67.8&69.2&70.0&34.6&57.1&48.4&73.8&58.3
			\\
			Retina300+att&ResNet101&61.5&76.5&70.6&66.2&42.1&34.1&69.3&59.4&87.2&42.6&59.0&47.5&83.8&69.0&72.9&71.6&37.3&58.9&48.3&74.7&59.9
			\\
			SSD300*&VGG16&75.8&88.1&82.9&74.4&61.9&47.6&82.7&78.8&91.5&58.1&80.0&64.1&89.4&85.7&85.5&82.6&50.2&79.8&73.6&86.6&72.1\\
			SSD321&ResNet101&75.4&87.9&82.9&73.7&61.5&45.3&81.4&75.6&92.6&57.4&78.3&65.0&90.8&86.8&85.8&81.5&50.3&78.1&75.3&85.2&72.5\\ 
			DSSD321&ResNet101&76.3&87.3&83.3&75.4&64.6&46.8&82.7&76.5&92.9&59.5&78.3&64.3&91.5&86.6&86.6&82.1&53.3&79.6&\textbf{75.7}&85.2&73.9\\
			ASSD300&VGG16&77.5&88.7&85.6&78.0&65.7&54.1&82.6&78.2&91.8&59.7&84.0&65.0&90.4&87.6&88.3&83.7&53.5&81.1&70.4&86.8&75.5		\\
			ASSD321 &ResNet101 &76.4&89.6&84.3&76.7&64.40&49.30&81.7&77.0&92.2&57.80&81.3&64.0&91.6&86.5&85.8&82.1&53.0&80.0&70.9&87.2&71.8
			\\
			\hline
			Retina512&ResNet101&67.7&80.4&74.0&73.4&53.5&49.7&73.0&71.2&88.2&45.8&69.7&50.6&87.1&74.0&76.8&78.9&45.6&69.1&51.3&77.2&65.0
			\\
			Retina512+att&ResNet101&68.8&81.4&77.6&73.3&54.1&53.0&74.3&72.27&85.01&48.5&71.5&50.0&87.6&77.4&77.3&80.0&49.5&71.6&53.2&72.9&66.3
			\\
			SSD512*&VGG16&78.5&90.0&85.3&77.7&64.3&58.5&85.1&84.3&92.6&61.3&83.4&65.1&89.9&88.5&88.2&85.5&54.4&82.4&70.7&87.1&75.6\\ 
			SSD513&ResNet101&79.4&90.7&87.3&78.3&66.3&56.5&84.1&83.7&94.2&62.9&84.5&66.3&92.9&88.6&87.9&85.7&55.1&83.6&74.3&88.2&76.8\\
			DSSD513&ResNet101&80.0&\textbf{92.1}&86.6&80.3&68.7&58.2&84.3&\textbf{85.0}&\textbf{94.6}&63.3&85.9&65.6&\textbf{93.0}&88.5&87.8&86.4&57.4&85.2&73.4&87.8&76.8\\
			ASSD512 &VGG16 & 80.0 &89.8&87.7&81.5&70.6&60.0&85.3&84.7&93.6&61.8&84.9&66.1&90.9&88.6&87.9&86.6&57.7&86.7&71.5&86.5&\textbf{77.4}
			\\
			ASSD513&ResNet101&\textbf{81.3}&92.1&\textbf{89.2}&\textbf{82.5}&\textbf{71.5}&60.4&\textbf{85.5}&84.8&93.9&\textbf{63.7}&\textbf{88.6}&\textbf{67.4}&92.6&\textbf{90.2}&\textbf{89.0}&\textbf{86.5}&\textbf{60.4}&\textbf{88.2}&73.4&\textbf{88.6}&77.0
			\\
			\hline
		\end{tabular}%
	}
	\label{tab:VOC2012}
\end{table*}
\begin{table*}[!h]
\caption{COCO test-dev detection results, which is evaluated by online evaluation server. We compared the our ASSD321 and ASSD512 performances with Faster R-CNN\cite{ren2015faster}, R-FCN\cite{dai2016r}, YOLOv2\cite{redmon2017yolo9000}, SSD300*, SSD500*\cite{liu2016ssd}, SSD321, SSD513, DSSD321, DSSD513 \cite{fu2017dssd}, FSSD300, FSSD512\cite{li2017fssd}, RetinaNet500\cite{lin2018focal}, RefineDet\cite{zhang2018single}.}
\centering
\resizebox{\textwidth}{!}{%
\begin{tabular}{|l|c|c|ccc|ccc|ccc|ccc|}\hline
\multirow{2}{*}{Method} & \multirow{2}{*}{Training Data}& \multirow{2}{*}{Backbone} & \multicolumn{3}{|c|}{Avg. Precision, IoU:} & \multicolumn{3}{|c|}{Avg. Precision, Area:} & \multicolumn{3}{|c|}{Avg. Recall, \#Dets:} & \multicolumn{3}{|c}{Avg. Recall, Area:} \\ 
 & & & 0.5:0.95 & 0.5 & 0.75 & S & M & L & 1 & 10 & 100 & S & M & L \\ \hline
 Faster R-CNN  & trainval&VGG16 &21.9&42.7&-& -&-&-& -&-&-& -&-&-
 \\
 R-FCN & trainval& ResNet101&29.9&51.9&-&10.8&32.8&45.0&-&-&-&-&-&-
 \\ 
 YOLOv2 &trainval35k&Darknet19&21.6&44.0&19.2&5.0&22.4&35.5&20.7&31.6&33.3&9.8&36.5&54.4\\
 \hline
 SSD300*& trainval35k& VGG16&25.1&43.1&25.8&6.6&25.9&41.4&23.7&35.1&37.2&11.2&40.4&58.4\\
SSD321 & trainval35k&ResNet101&28.0&45.4&29.3&6.2&28.3&49.3&25.9&37.8&39.9&11.5&43.3&64.9\\
 FSSD300& trainval35k& VGG16&27.1 &47.7& 27.8 &8.7& 29.2& 42.2 &24.6& 37.4& 40.0 &15.9& 44.2& 58.6\\
DSSD321& trainval35k&ResNet101&28.0&46.1&29.2&7.4&28.1&47.6&25.5&37.1&39.4&12.7&42.0&62.6\\
\textbf{ASSD321}& trainval35k&ResNet101&29.2&47.8&30.9&6.9&33.3&47.9&26.3&38.7&40.2&10.4&46.0&64.8\\
RefineDet320&trainval35k&VGG16&29.4& 49.2& 31.3 &10.0& 32.0& 44.4&-&-&-&-&-&-\\
\hline
 SSD512* &   trainval35k&VGG16&28.8&48.5&30.3&10.9&31.8&43.5&26.1&39.5&42.0&16.5&46.6&60.8\\
SSD513 & trainval35k&ResNet101&31.2&50.4&33.3&10.2&34.5&49.8&28.3&42.1&44.4&17.6&49.2&65.8\\
 FSSD512&trainval35k&  VGG16 & 31.8 &52.8& 33.5& 14.2& 35.1& 45.0& 27.6& 42.4& 45.0 &22.3& 49.9& 62.0  \\
DSSD513& trainval35k&ResNet101&33.2&53.3&35.2&13.0&35.4&51.1&28.9&43.5&46.2&21.8&49.1&66.4\\ 
RetinaNet500&trainval35k& ResNet101 & 34.4& 53.1& \textbf{36.8} &14.7& 38.5 &49.1& -&-&-& -&-&-
 \\
\textbf{ASSD513}& trainval35k&ResNet101&\textbf{34.5}&\textbf{55.5}&36.6&15.4&39.2&51.0&29.9&45.6&47.6&22.8&52.2&67.9\\
RefineDet512 &trainval35k&VGG16&33.0& 54.5&  35.5&  16.3&  36.3&  44.3&  -&-&-& -&-&-
 \\
\hline
\end{tabular}}
\label{tab:COCO}
\end{table*}

\subsection{Semantic Fusion} 
Motivated by FSSD \cite{li2017fssd}, we fuse the contextual information from layer4 and layer5 into layer3 to enrich its semantics. In our experiment, we find the fusion operation alone does not notably improve the detection accuracy (see Table~\ref{tab:Ablation}). Instead, it even decreases the accuracy a bit with more computational cost. The reason would be that the three layers possess different receptive fields and have different capabilities; further, the concatenation and $ 1\times1 $ conv transformation would possibly neutralize the relative importance of the three layers and suppress the critical features in original layer3. However, when we place the attention unit after the fusion operation, there is a noticeable improvement (see Table~\ref{tab:Ablation}). It is possible that semantics from the deep layers help the attention unit to discover useful information that resides in the original layer3. Finally, when only applying the attention unit, we observe inferior performance in contrast to the model with both fusion and attention mechanisms. This indicates that the feature fusion and attention are complementary to each other. The semantic fusion process can be formulated as:
\begin{equation}
\mathbf{x^3} = \mathbf{W^3}Concat\{\mathbf{x^3},\mathbf{x^4},\mathbf{x^5}\}+\mathbf{b^3},
\end{equation}
where $ \mathbf{x^s}\in \mathbb{R}^{C^s\times N^s} $ is the feature map at layer $ s $, $ \mathbf{W^3}\in \mathbb{R}^{C^3\times C^{\prime}} $ and $ \mathbf{b^3}\in \mathbb{R}^{C^3} $. In the concatenation operation, layer4 and layer5 are upsampled through bilinear interpolation in order to align their sizes with that of layer3.

\begin{figure*}[!h]
	\centering
	\includegraphics[width=1\textwidth]{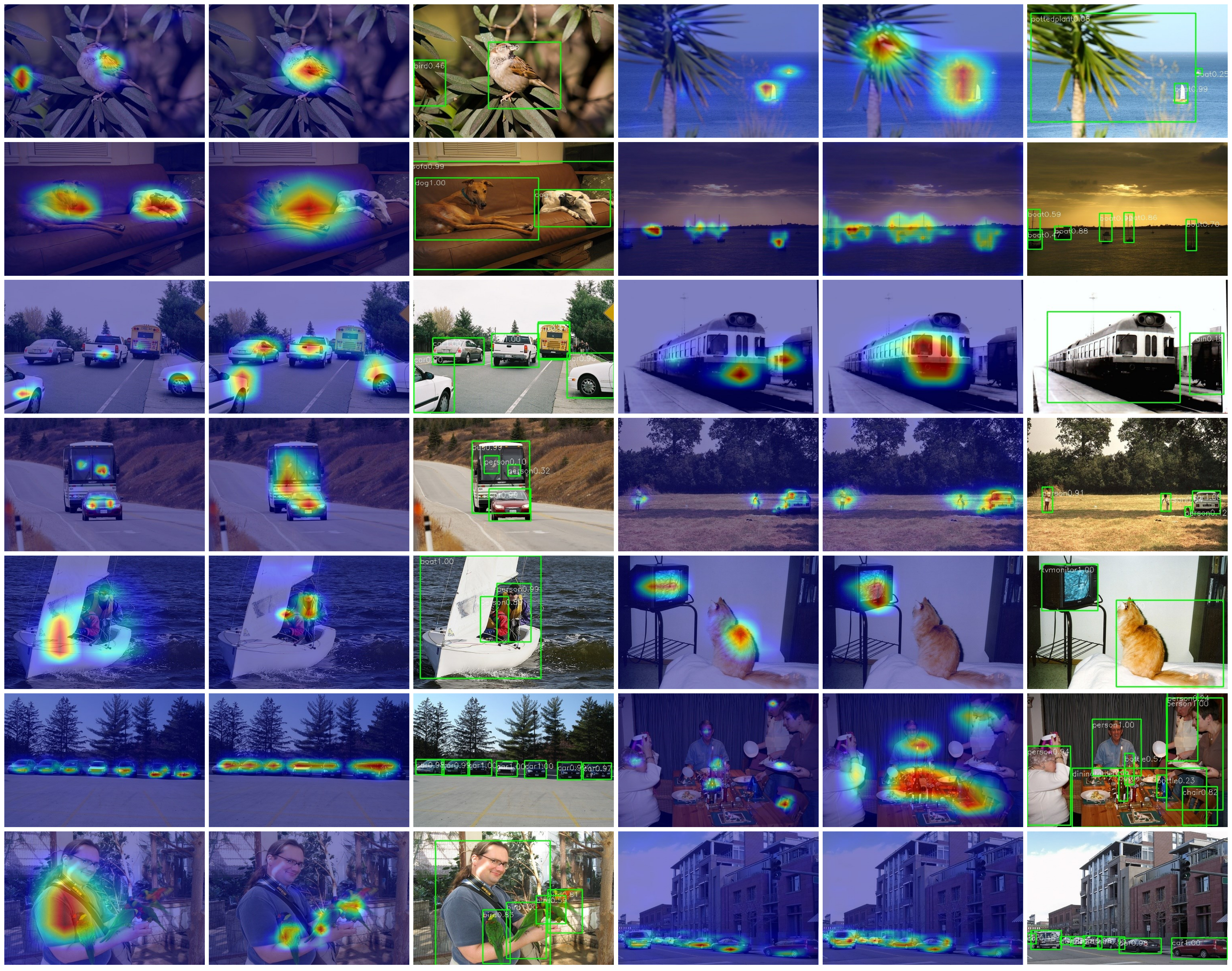}
	\caption{Visualization of attention maps on PASCAL VOC 2007 test set. The attention maps are calculated from feature maps of different scales. For a given input image, the attention maps highlight the useful regions of different sizes, as indicated by the heat regions. The attention map will be used as the weighted sum of spatial features at each location. Therefore, the features of unrelated regions such as background are suppressed. 
	In this way, the attention maps helps the model focus on the real targets and thereby improves the detection accuracy.
	}
	\label{fig:fig3}
\end{figure*}

\section{Implementation Details}
We follow the same anchor box generating method as SSD \cite{liu2016ssd}. Specifically, we use aspect ratio $a_r = \{1,2,1/2\} $ for anchor boxes on feature maps conv3,8,9 and $a_r=\{1,2,1/2,3,1/3\}$ for anchor boxes on feature maps of conv4-7. Each box has a minimum scale $s_{\min}$ and a maximum scale $s_{\max}$, where the scale $s_{\min}$ is regularly spaced over the feature map layers and $s_{\max}$ is the $s_{\min}$ of next layer. The normalized width and height of an anchor box are calculated by $w=s\sqrt{a_r}$ and $h=s/\sqrt{a_r}$, where $s=\sqrt{s_{\min}s_{\max}}$ for $a_r=1$, otherwise $s=s_{\min}$. We use hard negative mining to solve the positive-negative box class imbalance problem as in the original SSD \cite{liu2016ssd}. Also, we employ the same data augmentations and the same loss functions as SSD.

Our model is implemented with Pytorch \cite{paszke2017automatic} and trained on 8 NVIDIA Tesla K80 GPUs. The weights of ResNet101 backbone are pretrained on ImageNet. We use Stochastic Gradient Descent (SGD) algorithm to optimize ASSD weights, with a momentum of 0.9, a decay of 0.0005 and an initial learning rate of 0.001. Following the settings of SSD, DSSD and FSSD, we train and evaluate ASSD on two input resolution images: $321\times 321$ and $513\times 513$. In particular, we set the mini-batch size to 10 images per GPU for ASSD321 and 8 images per GPU for ASSD512.

\section{Experiments}
We conduct experiments on two common datasets: PASCAL VOC \cite{Everingham15} and COCO \cite{lin2014microsoft}. The PASCAL VOC dataset contains 20 object classes for object detection challenge. We evaluate ASSD on the PASCAL VOC 2007/2012 test set. The COCO dataset includes 80 object categories. In this work, we use COCO 2017 dataset, which has the same train, validation and test images as COCO 2014. Hence we have a fair comparison with the state-of-the-art methods. Note that RetinaNet \cite{lin2018focal} does not have PASCAL VOC detection results. Therefore we only compare the accuracy and speed of RetinaNet on COCO dataset.

\subsection{PASCAL VOC 2007}
We first evaluate our ASSD on PASCAL VOC 2007 test set with a primary goal of comparing the speed and accuracy of ASSD with state-of-the-art methods. The training dataset we use here is a union of 2007 trainval and 2012 trainval. We train ASSD321 for 280 epochs, where the initial learning rate of 0.001 decreases by 0.1 at the 200th epoch and the 250th epoch. For ASSD513, we train for 180 epochs, with a learning rate decay of 0.1 at the 120th and 170th epochs. As shown in Table~\ref{tab:VOC2007}, with a comparable fast speed, ASSD achieves a large improvement in accuracy compared to SSD, DSSD, and FSSD.

\subsection{Ablation study on PASCAL VOC 2007}
We perform ablation study to explore the effects of attention unit and semantic fusion on detection accuracy and speed. Here we investigate four models, SSD513, SSD513+fusion, SSD513+att, SSD513+fusion+att, on the PASCAL VOC 2007 test set. It can be observed from Table~\ref{tab:Ablation} that the fusion module alone does not show noticeable accuracy improvement. On the contrary, it brings a little more computational overhead. In contrast, attention unit alone leads to a significant performance improvement. When combining the attention unit with the fusion module, we observe further boost of performance. We conjecture that the attention unit may have the ability to analyze the contextual semantics at different levels and select the useful information for guiding a better detection.

\subsection{PASCAL VOC 2012}
We compare the detection accuracy of ASSD with the state-of-the-art methods on the PASCAL VOC 2012 test set. The mAP is evaluated by online PASCAL VOC evaluation server.  We present a detailed comparison of average precision (AP) for each class in Table~\ref{tab:VOC2012}. The training dataset contains 2007 trainval+test and 2012 trainval. We follow similar training settings as PASCAL VOC 2007. From Table~\ref{tab:VOC2012}, it can be seen that ASSD513 improves the detection accuracy for most of the classes. The reason would be that the attention unit figures out the pixel-level feature relationships and therefore enhances the model ability to distinguish objects of different classes.

\subsection{COCO}
We train and validate ASSD on COCO training dataset (118k) and validation dataset (5k). We compare with the state-of-the-art methods on COCO test-dev. The detection performance is evaluated by the online evaluation server. We train ASSD321 for 160 epochs with a learning rate decay of 0.1 at the 100th epoch and the 150th epoch. ASSD513 is trained for 140 epochs, and the learning rate decreases after 80 and 130 epochs. As illustrated in Table~\ref{tab:COCO}, ASSD achieves a large improvement over SSD, DSSD and FSSD. Besides, at a similar input resolution, ASSD513 obtains better accuracies than RetinaNet500, especially for AP at different object area thresholds. In particular, when the intersection over union (IoU) is higher than 0.5, ASSD513 has a 2.4\% improvement compared to RetinaNet500. Furthermore, from Table~\ref{tab:COCO} it can also be observed that ASSD is more effective at detecting the small, medium and large objects. Note that, with the above superiority in detection accuracy, ASSD513 (6.1FPS K40) still achieves comparable speed as RetinaNet500 (6.8FPS K40).

\subsection{Attention Visualization}
To better investigate the attention mechanism, we visualize the attention maps of different scales. In particular, we project the attention maps onto the original images. Here we utilize the PASCAL VOC 2007 test set, which contains 20 classes. From Fig.~\ref{fig:fig3}, we observe that the attention maps highlight the crucial locations of objects, indicating the feature relations help the model concentrate on useful regions. At shallow layers, the attention map guides the model to focus on small objects; while at deep layers, the attention map highlights objects with large sizes. Moreover, it can also be observed that the attention map suppresses the negative regions, which would be of great help for fast determination of negative anchor boxes.

\section{Conclusion}
In this paper, we propose an attentive single shot multibox detector, termed ASSD, for more effective object detection. Specifically, ASSD utilizes a fast and light-weight attention unit to help discover feature dependencies and focus the model on useful and relevant regions. ASSD improves the accuracy of SSD by a large margin at a small extra cost of computation. Moreover, ASSD competes favorably with the other state-of-the-art methods. In particular, it achieves better performance than the one-stage detector RetinaNet, while being easier to train without the need to heavily tune the loss parameters.


\end{document}